\documentclass[a4paper]{article}

\usepackage{INTERSPEECH2022}
\usepackage{graphicx}
\usepackage[caption=false]{subfig}
\usepackage{url}
\usepackage{enumitem}
\usepackage{multirow}
\usepackage{amsmath}
\usepackage{refcount}

\title{mcBERT: Momentum Contrastive Learning with BERT for Zero-Shot Slot Filling}

\name{Seong-Hwan Heo$^{\dagger\ast}$, WonKee Lee$^{\ddagger\ast}$\thanks{$^\ast$ Equal contribution to this work}, Jong-Hyeok Lee$^{\dagger \ddagger}$}
\address{
  $^\dagger$Graduate School of Artificial Intelligence,\\
  $^\ddagger$Department of Computer Science and Engineering, \\
  Pohang University of Science and Technology (POSTECH), Republic of Korea}
\email{\{hursung1, wklee, jhlee\}@postech.ac.kr}
\begin{document}

\maketitle
\begin{abstract}
Zero-shot slot filling has received considerable attention to cope with the problem of limited available data for the target domain.
One of the important factors in zero-shot learning is to make the model learn generalized and reliable representations. 
For this purpose, we present mcBERT, which stands for \textbf{`m'}omentum \textbf{`c'}ontrastive learning with \textbf{BERT}, to develop a robust zero-shot slot filling model.
mcBERT uses BERT to initialize the two encoders, the query encoder and key encoder, and is trained by applying momentum contrastive learning.
Our experimental results on the SNIPS benchmark show that mcBERT substantially outperforms the previous models, recording a new state-of-the-art.
Besides, we also show that each component composing mcBERT contributes to the performance improvement.
\end{abstract}

\section{Introduction}
Slot filling, an essential module in a goal-oriented dialog system, seeks to identify contiguous spans of words belonging to domain-specific slot types in a given user utterance.
For instance, given a user utterance \textit{``play some 70's music in youtube music''} belonging to the \texttt{Play music} domain\footnote{Sometimes `domain' is interchangeable with `intent'. However, we use the term `domain' throughout this paper.
}, the goal is to identify \textbf{slot entities}: \textit{`70's'} and \textit{``youtube music''} that correspond to the \textbf{slot types}, \texttt{year} and \texttt{service}, respectively.

Traditionally, slot filling models have relied on supervised learning by using labeled training data~\cite{DBLP:conf/interspeech/Young02, 10.1007/978-1-4614-8280-2_1, 6998838, kurata-etal-2016-leveraging}. Although having shown promising results for learned domains and slot types, these approaches require a significant amount of labeled data, which remains a chronic problem in developing robust systems.
Moreover, from a practical perspective, it is very natural to expect that any new domains (or new slot types), on which the model has not trained, can be issued to the dialog system.
Consequently, it is advisable for the model to maintain the ability to yield seamless predictions, even for domains and slot types that are rarely or even never learned.

\begin{figure}[t]
  \centering
  \includegraphics[width=0.8\linewidth]{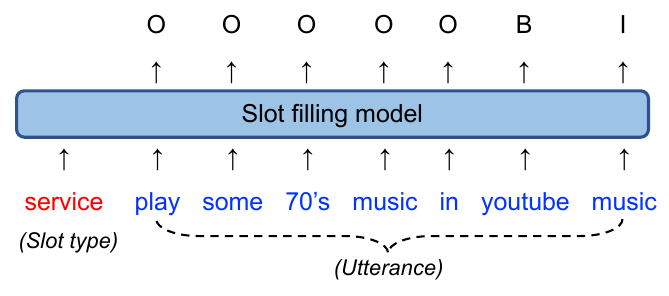}
  \caption{
        Schematic overview of slot filling under the cross domain framework.
        'B' and 'I' represent words corresponding to `service', whereas `O' represents a word not corresponding to `service'.
        }
  \label{fig1}
\end{figure}

In this regard, numerous recent studies focusing on \textbf{zero-shot} (and few-shot) slot filling have emerged to cope with limited training data. 
In particular, \textbf{cross domain framework}~\cite{46223, liu-etal-2020-coach, he-etal-2020-contrastive} has received much attention for zero-shot slot filling.
A basic concept of this framework is to leverage general knowledge across diverse domains when making predictions on a target domain, and for which a slot filling model is trained using the collection of labeled data from all available domains.
In addition, to enable zero-shot slot filling (especially to handle unseen slot types), a slot type and utterance are fed into the model simultaneously (Figure~\ref{fig1}) so that the model uses their semantic relationship (i.e., joint representation) to discover the slot entities corresponding to the given slot type.

\begin{figure*}[t]
    \centering
    \includegraphics[width=0.75\linewidth]{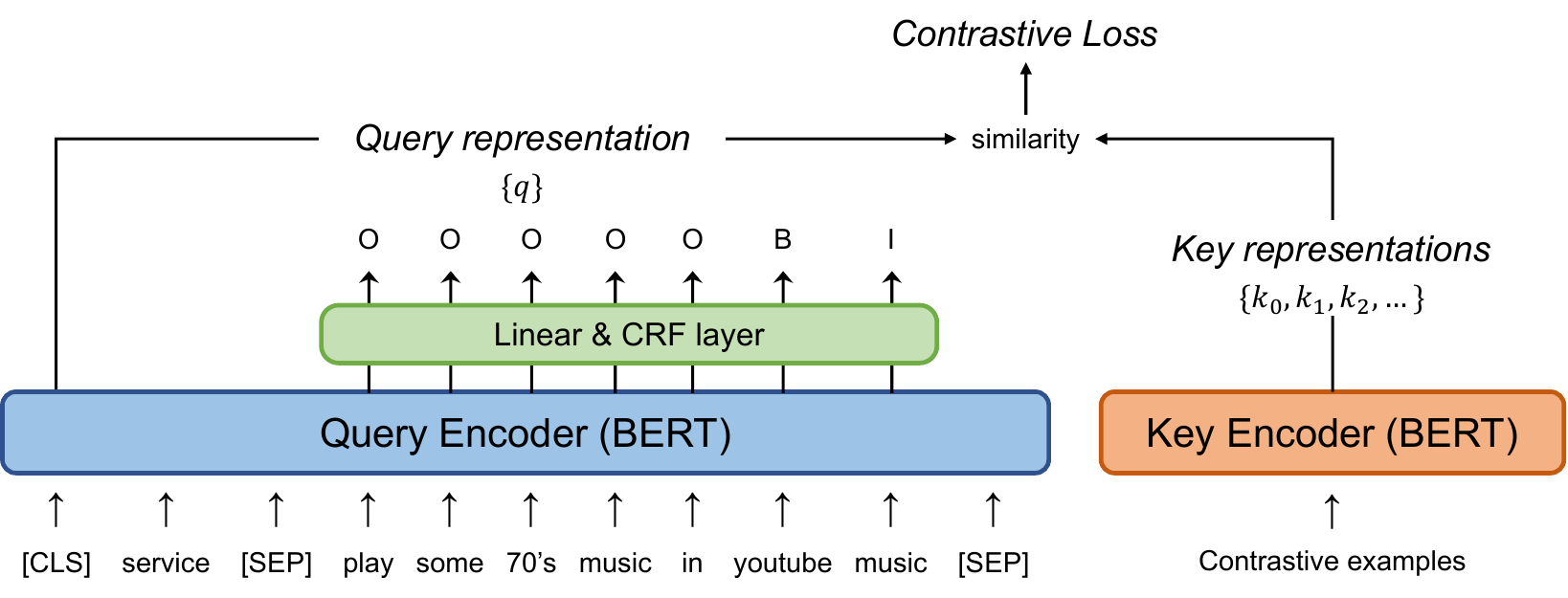}
    \caption{
    The overall architecture of mcBERT.}
    \label{fig2}
\end{figure*}

In recent years, great attention has been paid to \textbf{contrastive learning} for zero-shot learning in various research areas~\cite{9607851, 10.1145/3462244.3479904, xu-etal-2021-videoclip}, and several previous studies~\cite{liu-etal-2020-coach, he-etal-2020-contrastive} have also shown that contrastive learning is beneficial for zero-shot slot filling.
Basically, contrastive learning aims to construct contrastive samples, which include a single positive sample and several negative samples in response to the model input (also called anchor), and then let the model learn the representation of the anchor similar to the positive sample but dissimilar to all negative samples.
Particularly, in the computer vision field, \textbf{momentum contrast learning}~\cite{He_2020_CVPR} has achieved great success in learning unsupervised visual representations;
a key feature is to slowly update the \textbf{key encoder} (by which contrastive examples are encoded) using a momentum-based moving average of the \textbf{query encoder} (by which the anchor is encoded) to derive consistent representations between the anchor and contrastive samples.

In this paper, we present \textbf{`m'}omentum \textbf{`c'}ontrastive learning with \textbf{BERT} (\textbf{mcBERT}) for the zero-shot slot filling.
Beyond the visual representation, we hypothesize that momentum contrastive learning also has the potential to allow the model to learn adequate representations for zero-shot slot filling.
Moreover, to maximize the zero-shot effect, we utilize BERT \cite{devlin-etal-2019-bert}, a pre-trained language model with the capability of generating reliable and generalized linguistic representations, to initialize our query and key encoders.
Also, we consider two methods to construct contrastive samples by modifying the utterance:
(1) the first, inspired by~\cite{liu-etal-2020-coach}, is \textbf{template sample}, in which the slot entities in the utterance are replaced with slot types. 
(2) the second, inspired by~\cite{48487}, is \textbf{synthetic sample}, in which the slot entities in the utterance are replaced with different slot entities. 

Our experimental results on the SNIPS benchmark~\cite{DBLP:journals/corr/abs-1805-10190} revealed that mcBERT outperforms previous state-of-the-art models by a significant margin across all domains, both in zero-shot and few-shot settings, and we confirmed that each component we propose contributes to the performance improvement.

\section{Approach}
This section details our proposed method, mcBERT (Figure~\ref{fig2}).
First, we introduce the overall process of our BERT-based slot filling model, which operates under the cross-domain framework.
Next, we introduce a method to train our slot filling model by applying momentum contrastive learning, and for which we also introduce a method to construct contrastive samples.

\subsection{Slot Filling Model}
Slot filling is performed by using the \textbf{query encoder}'s outputs.
We use BERT as the query encoder because we expect that leveraging the knowledge transferred from BERT, which is a language model trained on a huge amount of plain text data, will significantly aid the model in learning proper representations for the slot filling task, especially for zero-shot slot filling.

Let $t$, $\mathbf{w}=(w_1, \cdots, w_n)$, and $\mathbf{y}=(y_1, \cdots, y_n)$ denote a slot type, an utterance including $n$ words, and a sequence of B/I/O labels corresponding to each $y_i$, respectively.
According to the BERT configuration, we configure the model input in the form of ``\texttt{[CLS]} $t$ \texttt{[SEP]} $\mathbf{w}$ \texttt{[SEP]}'' to accept both $t$ and $\mathbf{w}$.
After receiving this input, the query encoder outputs the hidden states $H$ with respect to $\mathbf{w}$, which also include information of $t$:
\begin{equation}
    H = [h_1, \cdots, h_n].
\end{equation}
Then, $H$ is forwarded to the linear projection layer to get the logits for the B/I/O labels:
\begin{equation}
    [\ell_1, \cdots, \ell_n] = \operatorname{Linear}(H).
\end{equation}
Subsequently, the CRF\footnote{To avoid clutter, we omit detailed calculation on CRF.} layer receives these logits to compute the label sequence score:
\begin{equation}
    score(\widetilde{\mathbf{y}}|\mathbf{w}) = \operatorname{CRF}\left([\ell_1, \cdots, \ell_n]\right).
\end{equation}
At the inference, the model selects a label sequence that maximize the score:
\begin{equation}
    \mathbf{y}^* = \mathrm{\underset{\widetilde{\mathbf{y}}}{\arg\max}}\left(score(\widetilde{\mathbf{y}} \mid \mathbf{w})\right).
\end{equation}
During training, the training object is to minimize the negative log likelihood with respect to the score as follows:
\begin{equation}
    \mathcal{L}_{BIO} = -\log \frac{score(\widetilde{\mathbf{y}} = \mathbf{y} \mid \mathbf{w})}{\sum_{\hat{\mathbf{y}}} score(\mathbf{\hat{y}} \mid \mathbf{w})} \label{bio_loss}
\end{equation}

\begin{figure}[t]
    \centering
    \subfloat[Template samples]{%
    \includegraphics[clip, width=\linewidth]{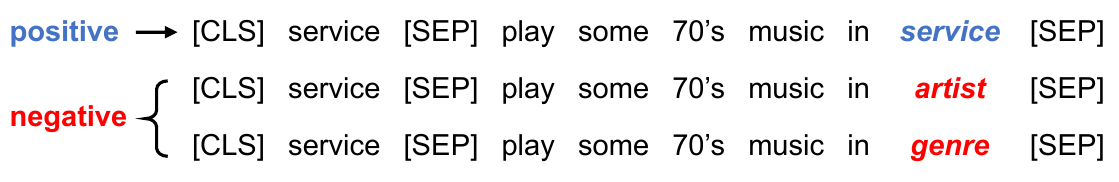}\label{fig3-1}%
    }
    
    \subfloat[Synthetic samples]{%
    \includegraphics[clip, width=\linewidth]{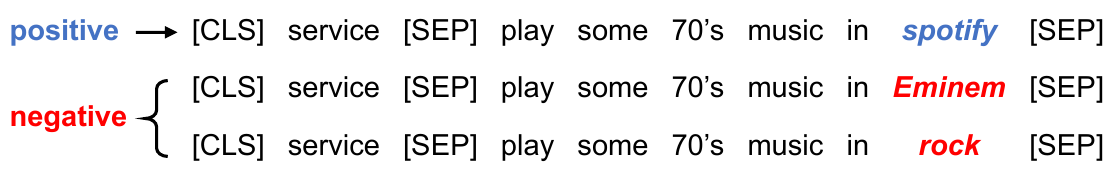}\label{fig3-2}%
    }
    \caption{Two types of contrastive samples given the anchor ``\texttt{[CLS]} service \texttt{[SEP]} play some 70's music in youtube music \texttt{[SEP]}'', whose slot entities are ``youtube music''.}
    \label{fig3}
\end{figure}

\subsection{Momentum Contrastive Learning}
\subsubsection{Contrastive sample construction} \label{sample construction}
Given the model input (i.e., anchor, which is the input to the query encoder), it is necessary to construct contrastive samples with size\footnote{Following previous studies~\cite{liu-etal-2020-coach, he-etal-2020-contrastive}, we set $m=3$ in our experiments.} $m$, containing a single \textit{positive} sample (similar to the anchor) and $m-1$ \textit{negative} samples (dissimilar to the anchor) so as to conduct contrastive learning.
We suggest two methods~(Figure~\ref{fig3}) for sample construction by modifying the given utterance.

The first approach, \textbf{template sample} (Figure~\ref{fig3-1}), replaces the slot entities in the utterance with a given slot type for the positive sample, but with a different slot type (randomly selected from data) for the negative samples.
This is to construct an utterance representation in which the representation of the slot type is encoded explicitly by lexicalizing the slot type into the utterance.
Thus, by mapping a given utterance to the positive sample yet straying from the negative samples, it is likely to create a representation of slot entities similar to its corresponding slot type.
Consequently, we expect that the model will more easily identify the slot entities corresponding to a given slot type.

The second approach, \textbf{synthetic sample} (Figure~\ref{fig3-2}), replaces the slot entities in the utterance with different slot entities (randomly selected from data), the slot types of which are still identical to a given slot type for the positive sample but different for the negative samples.
The intuition behind this approach is to construct positive samples containing different slot entities that are also likely to appear in a given utterance but to construct negative samples containing unnatural slot entities.
By using these samples in training, we expect the model to eventually better distinguish whether each word in an utterance is convincing regarding a given slot type.

To utilize both approaches, we empirically configure the contrastive samples by taking either template type or synthetic type at random with equal probabilities; we also analyze the effect on individual types and a different configuration, as detailed in Section~\ref{analysis}.

\subsubsection{Training process}
To perform contrastive learning, the \textbf{key encoder}\footnote{Note that the key encoder is not used for inference, only for training.} serving as a counterpart to the query encoder is necessary, being used to represent contrastive samples.
The key encoder is architecturally equivalent to the query encoder (i.e., the key encoder is also BERT) to make the representation of contrastive samples consistent with the anchor.

We use \texttt{[CLS]}, which encompasses all information on slot type and utterance, to construct the representation for both the anchor and contrastive samples.
Namely, the query encoder produces \textbf{query representation} $q$ from \texttt{[CLS]} in the anchor, and the key encoder produces \textbf{key representations} $K=\{k_1, \cdots, k_m\}$ from $m$ contrastive samples' \texttt{[CLS]};
the contrastive loss can be defined as
\begin{equation}
    \mathcal{L}_{CL} = -\log \frac{\textrm{exp}(q \cdot k_{+}/\tau)}{\sum_{k \in K} \mathrm{exp}(q \cdot k/\tau)},  \label{cl_loss}
\end{equation}
where $\tau$ is a temperature hyperparameter and $k_+$ is a single key regarding the positive sample that matches $q$.
In short, minimizing this loss is intended to classify $q$ as $k_+$.

Finally, apart from the key encoder, the model is trained to minimize the combined loss by using both $\mathcal{L}_{BIO}$ (Eq.~\ref{bio_loss}) and $\mathcal{L}_{CL}$ (Eq.~\ref{cl_loss}): 
\begin{equation}
    \mathcal{L} = \lambda\mathcal{L}_{CL} + (1-\lambda)\mathcal{L}_{BIO}, \label{combined_loss}
\end{equation}
where $\lambda \in [0, 1]$ is a hyperparameter to determine the ratio between two losses.
Following the original work~\cite{He_2020_CVPR}, the key encoder is trained with a momentum update as follows:
\begin{equation}
    \theta_k \leftarrow m\theta_k + (1-m) \theta_q, \label{mom_update}
\end{equation}
where $m\in[0, 1)$ is a momentum coefficient; $\theta_q$ and $\theta_k$ denote the parameters of the query encoder and key encoder, respectively.
This momentum update helps prevent $\theta_k$ from rapidly changing by keeping most of $\theta_k$ (the original work set $m=0.999$) while taking only a small portion of $\theta_q$ so that the key encoder can produce a consistent representation.

\begin{table*}[t]
\caption{
Experimental results on zero-shot and few-shot (with 50 samples) settings for the baselines and mcBERT in terms of F1-score.
The best results in each row (each domain) for zero-shot and few-shot learning are highlighted in \textbf{bold}.
The last rows are average F1-scores over all domains.
}
\label{results}
\centering
\begin{tabular}{l|ccccc|ccccc}
\toprule
\multirow{2}{*}{Domains}   & \multicolumn{5}{c|}{\textbf{Zero-shot}} & \multicolumn{5}{c}{\textbf{Few-shot} (50 samples)}                      \\ 
        & CT & RZT & Coach & \multicolumn{1}{c}{CZSL-Adv} & mcBERT & CT & RZT & Coach & \multicolumn{1}{c}{CZSL-Adv} & mcBERT \\ \midrule
AddToPlaylist      & 38.82 & 42.77 &50.90 & \multicolumn{1}{c}{53.89}    & \textbf{60.09}  & 68.69 & 74.89 & 74.68 & \multicolumn{1}{c}{76.18}    & \textbf{86.57}       \\ 
BookRestaurant     & 27.54 & 30.68 & 34.01 & \multicolumn{1}{c}{34.06}   & \textbf{71.64}  & 54.22 & 54.49 & 74.82 & \multicolumn{1}{c}{76.28}    & \textbf{84.82}       \\ 
GetWeather         & 46.45 & 50.28 & 50.47 & \multicolumn{1}{c}{52.24}   & \textbf{88.98}  & 63.23 & 58.87 & 79.64 & \multicolumn{1}{c}{83.28}    & \textbf{96.04}       \\
PlayMusic          & 32.86 & 33.12 & 32.01 & \multicolumn{1}{c}{34.59}   & \textbf{78.50}  & 54.32 & 59.20 & 66.38 & \multicolumn{1}{c}{68.17}    & \textbf{89.62}       \\ 
RateBook           & 14.54 & 16.43 & 22.06 & \multicolumn{1}{c}{31.53}   & \textbf{50.43}  & 76.45 & 76.87 & 84.62 & \multicolumn{1}{c}{87.22}    & \textbf{90.75}       \\ 
SearchCreativeWork & 39.79 & 44.45 & 46.65 & \multicolumn{1}{c}{50.61}   & \textbf{76.35}  & 66.38 & 67.81 & 64.56 & \multicolumn{1}{c}{66.49}    & \textbf{88.95}       \\ 
FindScreeningEvent & 13.83 & 12.25 & 25.63 & \multicolumn{1}{c}{30.05}   & \textbf{62.86}  & 70.67 & 74.58 & 83.85 & \multicolumn{1}{c}{83.26}    & \textbf{87.16}       \\ \midrule
Average F1         & 30.55 & 32.85 & \multicolumn{1}{c}{37.39} & \multicolumn{1}{c}{40.99}    & \multicolumn{1}{c|}{\textbf{69.84}} & 64.85 & 66.67 & \multicolumn{1}{c}{75.51} & \multicolumn{1}{c}{77.27}    & \multicolumn{1}{c}{\textbf{89.13}} \\ \bottomrule
\end{tabular}
\end{table*}

\begin{table}[h]
\caption{
Results on ablation study for each component used in mcBERT in terms of average F1-score.
The absence of BERT indicates that both encoders were trained from scratch.
The absence of $MoCo$ indicates that contrastive learning is performed with gradient updates instead of momentum updates.
}
\label{ablation}
\centering
\begin{tabular}{l|cc}
\toprule
Settings                            & Zero-shot & Few-shot                  \\ \midrule
mcBERT                                          & \textbf{69.84}    & \textbf{89.13}    \\
\ $-\ MoCo$             & 68.45             & 88.99    \\
\ $- \mathcal{L}_{CL} $                        & 66.21             & 88.62             \\
\ $-\ $ BERT                        & 35.25             & 59.22             \\ 
\midrule
\ $-\ $ BERT $-\ MoCo$  & 33.99             & 54.93             \\ 
\ $-\ $ BERT $- \mathcal{L}_{CL} $  & 35.92             & 60.02             \\ \bottomrule
\end{tabular}
\end{table}

\begin{table}[h]
\caption{
Comparison depending on the configuration of the contrastive samples.
Each result represents average F1-score.
}
\label{type_compare}
\centering
\begin{tabular}{l|cc}
\toprule
Contrastive samples        & Zero-shot & Few-shot \\ \midrule
Template  & 67.05 & \textbf{89.46}    \\
Synthetic  & 68.51 & 88.83   \\ 
Concat (Template, Synthetic)    & 68.29     & 88.94    \\ \midrule
Random (Template, Synthetic)    & \textbf{69.84}     & 89.13    \\ \bottomrule
\end{tabular}
\end{table}

\section{Experiments}
\paragraph*{Dataset.}
We used the SNIPS benchmark~\cite{DBLP:journals/corr/abs-1805-10190} to train and evaluate our model. 
SNIPS consists of 39 slot types across 7 domains, including approximately 2K samples per domain.
We tokenized all words in the dataset into sub-word units using the BERT tokenizer\footnote{\label{bert}\url{https://huggingface.co/bert-base-uncased}}.

\paragraph*{Model Configuration.}
We used the BERT model \texttt{bert-base-uncased}\footnotemark[\getrefnumber{bert}] to implement our query and key encoders.
For training, we used the following hyperparameters: the AdamW optimizer~\cite{loshchilov2018decoupled} with $\beta=(0.9, 0.999)$, learning rate of 1e-5, 4K warm-up steps followed by linear decay with 400K maximum training steps, a dropout probability of 30\%, and batch size of 128 samples.
We followed the original work~\cite{He_2020_CVPR} to configure the hyperparameters belonging to the momentum contrastive learning: $\tau=0.07$ (Eq.~\ref{cl_loss}) and $m=0.999$ (Eq.~\ref{mom_update}).

\paragraph*{Training Details.}
For a fair comparison, we followed the data setup described in \cite{liu-etal-2020-coach}, which has also been adopted by other baselines: 
(1) for each domain (i.e., the target domain) in SNIPS, the other six domains are selected as the source domains used for training;
(2) when conducting zero-shot learning, the data from the target domain are never used for training, 500 samples in the target domain are used for the development data, and the remainder are used as the test data; and
(3) when conducting few-shot learning, 50 samples from the target domain are used along with those from source domains for training; the development and test data configurations are the same as for zero-shot learning.
We trained mcBERT using a single NVIDIA RTX A5000 GPU, and it took approximately 2 hours to reach convergence.

\paragraph*{Baselines.}
In our experiments, we compared our mcBERT with the following baselines, including the current state-of-the-art system:
\begin{itemize}[leftmargin=*]
    \item \textbf{Concept Tagger (CT)} \cite{46223}: 
    a typical early work on cross-domain slot filling, where the RNN-based model receives both slot type and utterance simultaneously.
    
    \item \textbf{Robust Zero-shot Tagger (RZT)} \cite{48487}: 
    a method to improve CT, where examples of the slot entities are additionally provided to the model.
    
    \item \textbf{Coach} \cite{liu-etal-2020-coach}: 
    a method that equips two separate layers for detecting slot entities and slot type, respectively, and additionally provides template samples (similar to ours) to compute the regularization loss for model training.
    
    \item \textbf{CZSL-Adv} \cite{he-etal-2020-contrastive}: 
    current state-of-the-art (to our knowledge) model, which improved Coach by adopting contrastive learning (but not momentum contrastive learning) and by adding adversarial noise to the input.
\end{itemize}

\section{Results}
\subsection{Main Results}
The evaluation results (Table~\ref{results}) reveal that our proposed mcBERT is superior to all the baselines by a substantial margin in both zero-shot and few-shot settings for all domains, achieving a new state-of-the-art.
Note that the results for mcBERT used $\lambda=0.5$ for the zero-shot learning and $\lambda=0.1$ for the few-shot learning when configuring the combined loss (Eq.~\ref{combined_loss}), which resulted in the best performing model.
This indicates that momentum contrastive learning has a greater effect on zero-shot learning than few-shot learning.
We surmise that contrastive learning is less helpful in few-shot learning because the model can learn a suitable representation to some degree using at least a few data from the target domain.

\subsection{Analysis} \label{analysis}
We first conducted an ablation study to examine the effect of each component applied to mcBERT.
As shown in Table~\ref{ablation}, we confirmed that both BERT and momentum contrastive learning contributed to performance improvement.
The improvement gain was significant when BERT was used, which can be reasonable given that BERT has already been shown to be helpful in various tasks.
Furthermore, we observed that contrastive learning with momentum updates gave a greater performance enhancement than with gradient updates.
On the other hand, we found that contrastive learning was ineffective when training the encoders from scratch without using BERT, speculating that the given data were insufficient to train both the query and key encoder simultaneously.

Next, we investigated the effects of using individual types (i.e., template and synthetic types) for constructing contrastive samples and also using a different combination from that described in Section~\ref{sample construction}.
In Table~\ref{type_compare}, $\operatorname{Concat}(\cdot$) simply combines two types of samples, but the positive sample is randomly selected from one of the two types (as only a single positive sample is required); $\operatorname{Random}(\cdot$) is what we used to train mcBERT.
Consequently, we found that regardless of the configuration used, the performance was improved consistently; $\operatorname{Random}(\cdot$) achieved the best result for zero-shot learning, but using template type solely showed a slightly better result for few-shot compared to $\operatorname{Random}(\cdot$).

\section{Related Work}
Coach~\cite{liu-etal-2020-coach} introduced \textit{template regularization}, which can be conceptually similar to contrastive learning (but is not the same), and for which they constructed template examples in a similar manner to ours.
CZSL-Adv~\cite{he-etal-2020-contrastive} applied contrastive learning for training; their method differs from ours in that they did not apply momentum updates and their contrastive loss is calculated based on distance.

\section{Conclusions}
In this paper, we propose mcBERT especially focusing on zero-shot slot filling.
We attempt to apply momentum contrastive learning and use BERT to initialize our encoders.
To apply contrastive learning, we suggest constructing contrastive samples of two types: template and synthetic samples.
We found that our findings successfully contribute to the performance improvement and eventually achieved new a state-of-the-art.
This may stem from learning suitable representations, that is, we believe that our proposed model stably learns to cluster the slot types and their corresponding slot entities and captures semantic patterns precisely by momentum contrastive learning and BERT's representation.
We believe that our approach is applicable in various sequence labeling tasks, such as named entity recognition, which remains future work.


\section{Acknowledgements}
This work was supported by Institute of Information \& communications Technology Planning \& Evaluation (IITP) grant funded by the Korea government (MSIT) (No.2019-0-01906, Artificial Intelligence Graduate School Program (POSTECH)).

\bibliographystyle{IEEEtran}
\bibliography{mybib}
\end{document}